\documentclass[conference,a4paper]{IEEEtran}
\IEEEoverridecommandlockouts
\usepackage{cite}
\usepackage{amsmath,amssymb,amsfonts}
\usepackage{algorithm}
\usepackage{algpseudocode}
\usepackage{graphicx}
\usepackage{hyperref}
\usepackage{textcomp}
\usepackage{mathrsfs}
\usepackage{xcolor}
\def\BibTeX{{\rm B\kern-.05em{\sc i\kern-.025em b}\kern-.08em
    T\kern-.1667em\lower.7ex\hbox{E}\kern-.125emX}}
\begin{document}

\title{Regularizing Neural Network Training via Identity-wise Discriminative Feature Suppression} 

\author{\IEEEauthorblockN{Avraham Chapman}
\IEEEauthorblockA{\textit{School of Computer Science} \\
\textit{The University of Adelaide}\\
Adelaide, Australia \\
avraham.chapman@adelaide.edu.au}
\and
\IEEEauthorblockN{Lingqiao Liu}
\IEEEauthorblockA{\textit{School of Computer Science} \\
\textit{The University of Adelaide}\\
Adelaide, Australia \\
lingqiao.liu@adelaide.edu.au}
\thanks{Copyright 2022 IEEE. Published in the Digital Image Computing: Techniques and Applications, 2022 (DICTA 2022), 30 November – 2 December 2022 in Sydney, Australia. Personal use of this material is permitted. However, permission to reprint/republish this material for advertising or promotional purposes or for creating new collective works for resale or redistribution to servers or lists, or to reuse any copyrighted component of this work in other works, must be obtained from the IEEE. Contact: Manager, Copyrights and Permissions / IEEE Service Center / 445 Hoes Lane / P.O. Box 1331 / Piscataway, NJ 08855-1331, USA. Telephone: + Intl. 908-562-3966.}
}

\maketitle

\begin{abstract}
It is well-known that a deep neural network has a strong fitting capability and can easily achieve a low training error even with randomly assigned class labels. When the number of training samples is small, or the class labels are noisy, networks tend to memorize patterns specific to individual instances to minimize the training error. This leads to the issue of overfitting and poor generalisation performance. This paper explores a remedy by suppressing the network's tendency to rely on instance-specific patterns for empirical error minimisation. The proposed method is based on an adversarial training framework. It suppresses features that can be utilized to identify individual instances among samples within each class. This leads to classifiers only using features that are both discriminative across classes and common within each class. We call our method Adversarial Suppression of Identity Features (ASIF), and demonstrate the usefulness of this technique in boosting generalisation accuracy when faced with small datasets or noisy labels. Our source code is available.
\end{abstract}


\section{Introduction}
\label{section:intro}
In recent years, deep neural networks (DNNs) have grown in size from 3,246 trainable parameters in 1989 (LeNet \cite{LetNet}) to tens of millions of parameters (AlexNet \cite{AlexNet}, ResNet \cite{he2015resnet}). This has led to a massive increase in networks' ability to capture complex patterns from input data. However, this has also led to the risk of overfitting training data.
As an extreme case, Zhang et al. \cite{Overfitting} showed that any sufficiently large network can memorize the labels for instances in a dataset, even if they are randomly assigned. This effect is more pronounced when there is insufficient training data for a network or the labels associated with the data are noisy. The latter case is common when training data and their annotations are automatically crawled from the Internet.
 
Conceptually, an instance of training data could have two types of features:
\begin{itemize}
  \item \emph{Class-wise Discriminative Features}: These are features that are useful for determining the class that each sample belongs to. For example, a picture of a cat may be identified as such by the presence of whiskers.
  \item \emph{Identity-wise Discriminative Features}: These are features that are useful for determining precisely which sample is which. For example, the length and number of whiskers would help to identify a specific cat.
\end{itemize}
A deep neural network can use both features to perform empirical risk minimization. However, the second type of feature is less likely to generalise well to unseen data. Relying on it could cause overfitting.

Motivated by the above insight, this paper proposes a method to encourage the learning of Class-wise Features whilst discouraging the learning of Identity-wise Features. Intuitively, this leads to classifiers only using those features that are both discriminative across classes and common within each class. We do this by assigning each sample a unique ID and training a network that can classify samples whilst failing to determine the individual identity of each sample. This proposed process is performed adversarially in a manner akin to the DANN method \cite{ganin2015domainadversarial, ganin2015unsupervised} in domain adaptation. DANN maintains a domain classifier that can identify the domain of the input data while adversarially learning features that can reduce the discriminative power of the domain classifier. In our case, we essentially treat each individual sample as its own domain. 

In order to perform adversarial training, it was necessary to implement a gradient reversal layer (GRL) as described by Ganin et al. \cite{ganin2015domainadversarial}. However, proper adversarial training requires a balance between the regular backpropagation of the classification task and the reversed backpropagation of the sample identification task. This requires careful fine-tuning. While it is possible to discover a training schedule that maintains this balance, it is tricky and subject to error. We therefore also propose a method called Dynamic Gradient Reversal (DGR), which requires no tuning. Experimentation shows that DGR can be used anywhere a GRL is used.

We further explore two use cases for ASIF. The first one is to improve the generalisation performance of a deep neural network when training with a small amount of data. The second one is to increase the resilience against inaccurately labelled training data. In the second case, we show that the proposed method can be directly applied as a regularization approach or can be used to identify the incorrect class labels. 

In short, the main contributions of this paper are:

\begin{itemize}
  \item A technique to reduce the tendency for large networks to overfit to specific samples in the training data.
  \item A training method that maintains or improves accuracy while reducing the number of class-variant features to the absolute minimum.
  \item A gradient reversal algorithm that can be used anywhere a DANN-style Gradient Reversal Layer is used.
\end{itemize}

The source code behind these experiments is available at \url{https://github.com/avichapman/identity-feature-suppression}.

The rest of the paper is organised as follows. In Sect. \ref{RelatedWorks}, we provide further background in the problem area. We then describe the methodology behind our approach in Sect. \ref{Method}. In Sect. \ref{Experiments}, we describe the experiments performed to explore the characteristics of ASIF. Finally, in Sect. \ref{Conclusions} we summarise our results and suggest fruitful directions for future work.

\section{Related Works}
\label{RelatedWorks}

\subsection{Domain Adaptation}

Domain Adaptation is a family of techniques for learning a task on data from one domain and applying that task to another domain \cite{Wilson2020Survey}. Pan and Yang \cite{PanSurvey2010} define a 'domain' as consisting of a feature space and a marginal probability distribution of the features across the population.

Domain Adaptation works on the assumption that the feature space stays the same. That is, the things about a data sample that can be measured never change. For example, in any given country it is possible to measure the probability of a pregnancy resulting in fraternal twins. However, the marginal probability distribution between domains can vary widely. Applied to the aforementioned example, this would mean that the probability of fraternal twins in one country may be very different from that same measurement in another country.

Several recent works \cite{Dredze2009MultidomainLB, joshi-etal-2012-multi, hassan2018unsupervised} have attempted to take advantage of data across all domains to learn to predict within a single domain. Sebag et al. \cite{schoenauersebag2019multidomain} attempted to use adversarial learning to learn the distributions of the various domains in the training set.

Wilson, Doppa and Cook \cite{Wilson2020} proposed a method called Convolutional deep Domain Adaptation model for Time Series data (CoDATS), which took advantage of adversarial learning to encourage a feature extractor network to learn features that were domain invariant. The CoDATS network consisted of three parts: a feature extractor, a classifier and a domain predictor. The classifier was trained against labelled data for a source domain. At the same time, the domain predictor was trained to predict whether a given sample belonged to the source domain or a target domain. They used a gradient reversal layer (DGR) to create an adversarial relationship between the two tasks. The DGR worked by passing a feature vector through untouched when feeding forward and reversing the gradient when backpropagating.

Using this technique, Wilson et al. were able to demonstrate superior classification on the target domain, despite the network having never seen labels for the target domain. However, this technique is limited in that it requires each domain to have a label, which doesn't work in blurry edge cases or when the target domain is unknown at training time.

\subsection{Learning from Noisy Labels}
\label{section:RelatedWorksNoisyLabels}

In real life datasets, perfect label accuracy is unrealistic. Human annotators make errors due to fatigue and other considerations. Moreover, labels are often obtained through means such as Amazon's Mechanical Turk or as pseudo-labels generated via semi-supervised means.

Referred to as \emph{noisy labels}, these inaccurate labels can have a deleterious effect on training accuracy. Zhang et al. \cite{DBLP:journals/corr/ZhangBHRV16} demonstrated that a sufficiently complex DNN could learn a dataset with an arbitrarily high level of label noise. This overfitting behaviour negatively affects performance when evaluated on test data.

Fr\'enay and Verleysen \cite{FrenayLabelNoiseSurvey} divided label noise into two types: \emph{Instance-independent} and \emph{Instance-dependent}.

\emph{Instance-independent} label noise is characterised by the lack of a probabilistic relationship between a label being wrong and the underlying features of a given sample. In the literature, this is often further sub-divided into Symmetric and Asymmetric noise.

\begin{itemize}
  \item Symmetric noisy labels are modelled by randomly changing label values from their true class to some other class with a certain probability.
  \item Asymmetric label noise is modelled by randomly changing the labels for samples of certain classes to a similar class with a given probability. An example of this would be changing 'cars' to 'trucks' in the CIFAR10 dataset. This is slightly more realistic.
\end{itemize}

\emph{Instance-dependent} label noise, on the other hand, varies with the particular characteristics of each sample. This mirrors reality, in that more ambiguous samples are more likely to be misclassified. 

Chen et al. \cite{Chen2021InstanceDependentNoise} described a method for producing realistic instance-dependent noise. Their method involved training a DNN on a dataset for a certain number of epochs and recording the associated loss of each sample. The samples with the highest averaged loss were deemed to be the most counter-intuitive samples and therefore the most likely to be mislabelled in real life. While this does not work as an online means of determining misclassification likelihood, it is more than sufficient when the training data is known beforehand.

DNNs tend to learn general classification rules first before proceeding to memorise individual samples in a dataset. This means that early in training, a sample's classification loss can be used as an indication of whether the sample's label is incorrect. A large loss may indicate a label is wrong. This is often referred to as the \emph{small-loss trick}. Han et al. \cite{Han2018CoTeaching}, Jiang et al. \cite{Jiang2018Mentornet} and Yu et al. \cite{Yu2019Disagreement} all utilise the small-loss trick to detect noisy labels.

A popular method for combating overfitting to noisy labels is to change the rate at which DNNs learn instance-specific cases by modifying the loss function. This means that samples with an abnormally high loss (and thus likely to be wrongly labelled) are ignored or have reduced affect on the training outcome. Zhang et al. \cite{GCE} proposed the Generalised Cross Entropy loss, which bridged the gap between Cross Entropy loss and the MAE/unhinged loss. Menon et al. \cite{PHUber} went further and proposed a partially Huberised Cross Entropy loss, which utilized gradient clipping to arrive at a more robust training solution. Both of these methods are included for comparison in this paper.

\section{Method}
\label{Method}

This section describes the proposed ASIF training method in further detail.

\subsection{Notations and Definitions}

The following notation will be used in this paper:

\begin{itemize}
  \item $N$: The total number of samples being trained on.
  \item $C$: The total number of classes in the dataset.
  \item $N_c$: The total number of samples of a given class.
  \item $B$: The total number of samples in each mini-batch.
  \item $\eta$: The amount of label noise, 0-1.
  \item $\mathscr{L}_{\text{id}_c}$: The identification task loss associated with class $c$.
  \item $\mathscr{L}_\text{cls}$: The classification task loss.
  \item $\Psi$: The Feature Extractor.
  \item $F_\Psi$: The dimension of the feature vector output from $\Psi$.
  \item $h_I$: The Identifier module.
  \item $h_C$: The Classifier module.
\end{itemize}

\subsection{Method Description}

As discussed in Section \ref{section:intro}, ASIF attempts to overcome the problem of networks overfitting to specific instances in the training data. To that end, it attempts to select \emph{Class-wise Features} while suppressing \emph{Identity-wise Features}. Recall that the former are features useful in classifying a sample into one of several classes, while the latter features are useful in determining the identity of a specific sample.

ASIF attempts to perform both tasks: \emph{Classification} and \emph{Identification}. \emph{Classification} divides the dataset into $C$ classes and attempts to determine the class that each sample belongs to. Similarly, \emph{Identification} divides the dataset into $N$ 'identities' (one for each sample) and attempts to ascertain the identity of each and every sample. The module contains global parameters as well as parameters that are tuned for each class to perform identification amongst the individual samples within that class. A gradient reversal layer exists to encourage the network to fail in its \emph{Identification} task.

\subsection{Network Structure}

\begin{figure*}
\centering
    \includegraphics[width=\textwidth]{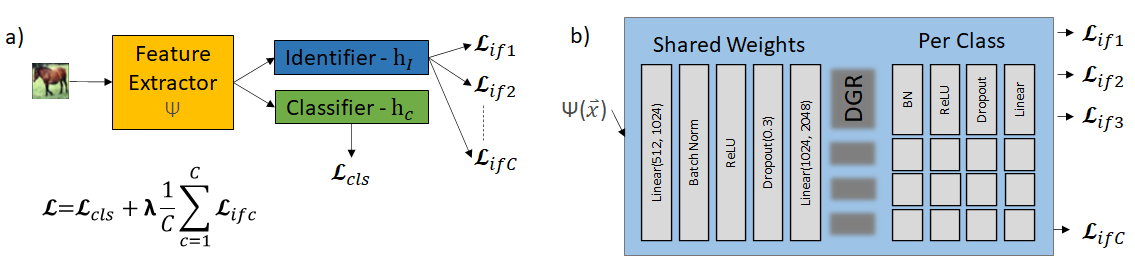}
    \caption{Architecture of our proposed ASIF network. \emph{(a)} A shared Feature Extractor $\Psi$ extracts features for use by all downstream tasks. A Classifier module $h_C$ performs the \emph{Classification} task, while an Identifier module $h_I$ performs the \emph{Identification} task. \emph{(b)} The Identifier module contains shared parameters that are trained on all samples, as well as dedicated parameters for each class of samples producing $C$ outputs, one for each class. There is a Dynamic Gradient Reversal (DGR) layer between the shared and per-class parameters.}
    \label{fig:asif_model}
\end{figure*}

As Figure \ref{fig:asif_model} indicates, the ASIF network consists of several components:

\subsubsection{Feature Extractor}

The Feature Extractor ($\Psi$) can be any off-the-shelf network. For the purposes of the experiments we've done as part of this paper, we used ResNet18 \cite{he2015resnet}. For ease of base-lining, we used the ResNet18 implementation provided as part of Nishi et al.'s \cite{Nishi2021Augmentation} work.

The output of the feature extractor is a feature vector $\Psi(x)$, where $\Psi(x) \in \mathbf{R}^{F_\Psi}$. This is passed to a linear layer $h_C$, the 'Classifier' in Figure \ref{fig:asif_model}(a), to extract logits for use with the \emph{Classification} task. The logits are combined with the classification label using a Cross Entropy loss ($\mathscr{L}_\text{cls}$). Note that the loss applied here can be varied. Investigation of other losses for the Classification task is left to future work.

\subsubsection{Identification Module}

The Identification Module $h_I$ attempts to identify individual samples. It has $C$ outputs,
one for each class, where $h_{I_c}(\Psi(x)) \in \mathbf{R}^{N_c}$.

As shown in Figure \ref{fig:asif_model}(b), the module is divided into three parts: a 'public' part whose parameters are trained on all samples regardless of label, a DGR, and $C$ sets of private parameters that are only trained on samples with the matching label. Each set of private parameters $1..C$ outputs one of the outputs $h_{I_c}(\Psi(x))$.

By sharing as many parameters as possible between the classes, we allow for a network that more easily scales to a larger number of classes.

The 'public' section of the Identifier consists of two linear layers with a Batch Normalization, ReLU and dropout in between. The 'private' section has a Batch Normalization, ReLU, dropout and final linear layer for each class. All hyperparameters, such as dropout levels, were manually optimised using cross-validation.

Each mini-batch output from the Identifier is filtered down to just the batch members from its class and a Cross-Entropy Loss ($\mathscr{L}_\text{id}$) is applied. The losses are then averaged in proportion to each class' share of the mini batch and combined with the classification loss. This structure allows the ASIF network to optimise:

\begin{equation}
\label{eqn:asif_loss}
\min_{\Psi, h_C}\max_{\{{h_I}_c\}}\mathbb{E}[h_C(\Psi(x))] \\ + \lambda_\text{id}\mathbb{E}[\frac{1}{C}\sum\limits_{c=1}^{C} {h_I}_c(\Psi(x))],
\end{equation}
where ${h_I}_c$ denotes the Identifier output for class $c$.
It also includes a coefficient $\lambda_\text{id}$ for the Identifier losses. The values used in the experiments can be found in the appendix.

\subsubsection{Dynamic Gradient Reversal Layers}

Similar to DANN \cite{ganin2015domainadversarial}, we reverse the gradient during backpropagation from the \emph{Identification} task.
To review, DANN contains a Gradient Reversal Layer (GRL) which, on feeding forward has the value $R(x) = x$. However, when backpropagating it has the value $\frac{\partial R}{\partial x} = -\lambda I$, where $I$ is the identity matrix and $\lambda$ is a hyperparameter. The value of $\lambda$ is set according to a schedule tuned to ensure a proper balance is maintained between the competing tasks. Unfortunately, choosing an incorrect value for $\lambda$ leads to the \emph{Identification} task becoming confidently wrong.

\begin{figure}
    \centering
    \includegraphics[width=3.2in]{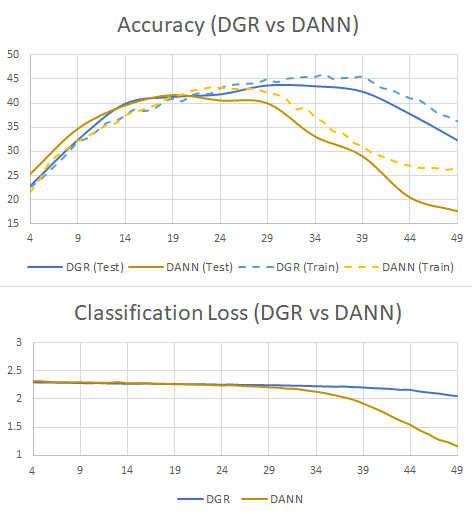}
    \caption{Classification loss, training and test accuracy versus training epoch when training with Symmetrical 80\% noise and CIFAR10. The use of Dynamic Gradient Reversal (DGR) leads to reduced overfitting, as indicated by the classification loss not dropping as fast. Also note that the DGR training and test accuracies remain in lockstep, while the DANN accuracies diverge during later training.}
    \label{fig:dgr_vs_dann}
\end{figure}

To overcome this limitation, this paper proposes a Dynamic Gradient Reversal (DGR) scheme. Unlike DANN, our method does not require a tune-able hyper-parameter. Moreover, our experiment shows that DGR alleviates overfitting better when compared with DANN-like gradient reversal layers, as shown in Figure \ref{fig:dgr_vs_dann}.

\begin{algorithm}
\caption{Dynamic Gradient Reversal}\label{alg:dgr}
\begin{algorithmic}
\State $\hat{\mathscr{L}_\text{id}} \gets log(N_c)$
\State $\lambda \gets 1.0$
\While{Still Training}
    \State Perform \emph{Identification Task}
    \State Backpropagate using $\lambda$
    \State Record $\mathscr{L}_\text{id}$ 
    \State $\lambda \gets \frac{\mathscr{L}_\text{id} - \hat{\mathscr{L}_\text{id}}}{\hat{\mathscr{L}_\text{id}}}$
\EndWhile
\end{algorithmic}
\end{algorithm}

Formally, Dynamic Gradient Reversal (DGR) is designed to maintain the \emph{Identification Task} in a state of maximum uncertainty. This is done by establishing a baseline ideal loss ($\hat{\mathscr{L}_\text{id}}$) corresponding to maximum entropy in the identity logits as shown in Equation \ref{eqn:max_uncertainty}:

\begin{equation}
\label{eqn:max_uncertainty}
\hat{\mathscr{L}_\text{id}} = -\sum_{I=1}^{N_c} \frac{1}{N_c} log(\frac{1}{N_c}) = log(N_c).
\end{equation}

Then we automatically choose $\lambda$ by comparing the current $\mathscr{L}_\text{id}$ and $\hat{\mathscr{L}_\text{id}}$.
The detailed process of calculating $\lambda$ is described in Algorithm \ref{alg:dgr}. The algorithm has the same computational cost as DANN, since the tuning of the $\lambda$ value is done using the loss, which is computed anyway.

\subsection{Noisy Label Detection}

One of the challenges of obtaining real-life data is the difficulty of creating accurate labels. It is typical to use a large human workforce and/or some degree of automation (for example, scraping the Internet) to acquire the labels. Inaccurate (or 'noisy') labels often become associated with the training data. This can lead to reduced accuracy in trained networks.

To detect noisy labels, the \emph{small-loss trick} \cite{Han2018CoTeaching, Jiang2018Mentornet, Yu2019Disagreement} can be applied, which is described above in Section  \ref{section:RelatedWorksNoisyLabels}. DNNs learn general cases first and high losses indicate tricky cases or bad labels. Given ASIF's regularising effect during training, the feature extractor will take much longer to overfit to the edge cases. We thus theorise that the small-loss trick applied to the Classification Task's loss $\mathscr{L}_\text{cls}$, would be far more robust when trained with ASIF than without.

\section{Experiments}
\label{Experiments}

In this section, we evaluate the performance of ASIF using reduced training sets and noisy labels. We also delve into the characteristics of ASIF. Experiments were run using both the CIFAR10 and Fashion-MNIST \cite{FashionMNIST} datasets. All training techniques are judged based on their macro F1 scores when classifying on the test set.

The CIFAR10 dataset contains 50,000 small 32x32 sample images, split evenly across ten classes. It contains a further 10,000 images for evaluation - 1000 per class. The Fashion-MNIST dataset contains 60,000 small 28x28 grayscale images of fashion products, split evenly across ten classes. It contains a further 10,000 images for evaluation.

To serve as a point of comparison, all experimental configurations were also run with several other methods. In these non-ASIF cases, the Identifier module was removed from the network. Three different losses were applied to the output logits from the Classifier:

\begin{itemize}
  \item CE: Cross-Entropy. This is the same as used in the ASIF experiments.
  \item GCE: Generalised Cross-Entropy \cite{GCE}
  \item PHuber: partially Huberised Cross-Entropy \cite{PHUber}
\end{itemize}

\subsection{Reduced Training Sets}
\label{ReducedTrainingSets}

To investigate ASIF's ability to resist overfitting with small training sets, we designed an experiment which trained the network on the CIFAR10 dataset with various numbers of samples per class.

The inputs to all techniques were the same, with standard regularisation applied, but no data augmentation. Runs were conducted with $N$ = [10k, 20k, 30k, 40k, all]. Each configuration was run three times and their macro F1 scores across the test set averaged.

\begin{table}[]
    \centering
\begin{tabular}{||c c c c c||} 
 \hline
 $N$ & CE & GCE & PHuber & ASIF \\ [0.ex] 
 \hline\hline
 10k & 69.1 \textpm \, 1.2 & 64.8 \textpm \, 2.4 & 73.2 \textpm \, 0.6 & \textbf{74.2 \textpm \, 0.2} \\ 
 \hline
 20k & 74.0 \textpm \, 0.5 & 72.6 \textpm \, 1.1 & 79.4 \textpm \, 0.0 & \textbf{80.2 \textpm \, 0.1} \\ 
  \hline
 30k & 78.7 \textpm \, 0.4 & 75.6 \textpm \, 0.7 & 82.1 \textpm \, 0.1 & \textbf{83.6 \textpm \, 0.3} \\ 
  \hline
 40k & 81.5 \textpm \, 0.7 & 80.6 \textpm \, 0.1 & 83.4 \textpm \, 0.1 & \textbf{85.5 \textpm \, 0.3} \\ 
  \hline
 50k & 84.3 \textpm \, 0.6 & 83.0 \textpm \, 0.4 & 83.9 \textpm \, 0.4 & \textbf{86.8 \textpm \, 0.4} \\ [1ex] 
  \hline
\end{tabular}
    \caption{Macro F1 Scores when training on reduced training sets on CIFAR10.}
    \label{tab:small_set_cifar10_results}
\end{table}

We discovered that ASIF has a marked advantage over its competitors in this domain. Table \ref{tab:small_set_cifar10_results} summarises the results for CIFAR10. Fashion-MNIST results can be found in Table \ref{tab:small_set_fashion_mnist_results} in the appendix.

\subsection{Training with Label Noise}

To test ASIF in the presence of label noise, we intentionally modified the labels in the datasets. Two types of label noise were investigated.

The first was Symmetrical instance-invariant noise, as described by Zhang et al. \cite{DBLP:journals/corr/ZhangBHRV16} and by Zhang and Sabuncu \cite{DBLP:journals/corr/abs-1805-07836}. To apply this sort of noise, a percentage of the dataset corresponding to the desired noise level $\eta$ were selected for label modification.

Symmetrical instance-invariant noise is not realistic, since labelling errors are more likely to happen with ambiguous samples than with obviously distinct ones. We used the technique described by Chen et al. \cite{Chen2021InstanceDependentNoise} to produce realistic instance-dependent noise. Their process produced a list of samples and associated average losses. We ranked the samples by descending loss and selected the top $N \times \eta$ samples. Those samples then had their labels randomly swapped.

\begin{figure}
    \centering
    \includegraphics[width=3.2in]{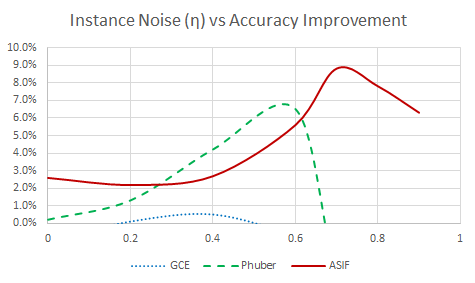}
    \caption{CIFAR10 results. ASIF confers a clear accuracy improvement over the baseline training method (Cross Entropy) with noisy labels, especially in the high-noise end of the range. GCE and PHuber training results are included for comparison.}
    \label{fig:inst_noise_vs_accuracy_improvement}
\end{figure}

Having trained ASIF against six values of $\eta$ and two different noise techniques, we have been able to show superior performance in the very high noise domain around 70\%-90\% noise. Figure \ref{fig:inst_noise_vs_accuracy_improvement} clearly illustrates ASIF's accuracy advantage in that area with CIFAR10.

In the easier scenario of Symmetric Instance-Invariant noise, PHuber is still competitive with ASIF. However, even in those cases, ASIF beats CE and GCE by a wide margin. Please see Tables \ref{tab:noise_cifar10_inst} and \ref{tab:noise_cifar10_sym} for CIFAR10 full results. ASIF's corresponding performance with Fashion-MNIST can be found in Tables \ref{tab:noise_fashion_mnist_sym} and \ref{tab:noise_fashion_mnist_inst} the appendix.

\begin{table}[]
    \centering
\begin{tabular}{||c c c c c||} 
 \hline
 $\eta$ & CE & GCE & PHuber & ASIF \\ [0.ex] 
 \hline\hline
 0 & 84.3 \textpm \, 0.6 & 83.0 \textpm \, 0.4 & 83.9 \textpm \, 0.4 & \textbf{86.8 \textpm \, 0.4} \\ 
 \hline
 0.2 & 79.5 \textpm \, 0.5 & 79.1 \textpm \, 0.1 & 80.5 \textpm \, 0.3 & \textbf{81.5 \textpm \, 0.0} \\ 
  \hline
 0.4 & 70.4 \textpm \, 0.3 & 70.5 \textpm \, 0.1 & \textbf{74.2 \textpm \, 0.4} & 73.1 \textpm \, 0.6 \\ 
 \hline
 0.6 & 59.2 \textpm \, 0.7 & 58.4 \textpm \, 0.3 & \textbf{65.7 \textpm \, 0.1} & 65.6 \textpm \, 1.9 \\ 
 \hline
 0.7 & 52.8 \textpm \, 1.0 & 50.9 \textpm \, 0.6 & 44.1 \textpm \, 0.6 & \textbf{61.6 \textpm \, 0.9} \\ 
 \hline
 0.8 & 45.3 \textpm \, 1.4 & 42.4 \textpm \, 0.5 & 34.0 \textpm \, 0.2 & \textbf{53.5 \textpm \, 1.0} \\
 \hline
 0.9 & 38.2 \textpm \, 0.5 & 31.5 \textpm \, 0.9 & 27.4 \textpm \, 2.3 & \textbf{41.8 \textpm \, 1.3} \\ [1ex] 
 \hline
\end{tabular}
    \caption{Macro F1 Scores with Instance-Dependent Noisy Labels on CIFAR10.}
    \label{tab:noise_cifar10_inst}
\end{table}

\begin{table}[]
    \centering
\begin{tabular}{||c c c c c||} 
 \hline
 $\eta$ & CE & GCE & PHuber & ASIF \\ [0.ex] 
 \hline\hline
 0 & 84.3 \textpm \, 0.6 & 83.0 \textpm \, 0.4 & 83.9 \textpm \, 0.4 & \textbf{86.8 \textpm \, 0.4} \\ 
 \hline
 0.2 & 65.5 \textpm \, 0.5 & 75.1 \textpm \, 0.4 & \textbf{81.8 \textpm \, 0.1} & 77.6 \textpm \, 0.7 \\ 
 \hline
 0.4 & 49.0  \textpm \, 2.3 & 60.3 \textpm \, 1.1 & \textbf{78.6 \textpm \, 0.8} & 72.0 \textpm \, 0.4 \\ 
 \hline
 0.6 & 41.0 \textpm \, 1.1 & 48.3 \textpm \, 1.3 & \textbf{73.5 \textpm \, 0.3} & 63.3 \textpm \, 1.9 \\ 
 \hline
 0.7 & 29.7 \textpm \, 1.1 & 40.8 \textpm \, 0.5 & \textbf{67.3 \textpm \, 1.3} & 55.6 \textpm \, 0.5 \\ 
 \hline
 0.8 & 21.4 \textpm \, 0.7 & 29.3 \textpm \, 0.9 & \textbf{53.6 \textpm \, 4.4} & 43.1 \textpm \, 2.4 \\
 \hline
 0.9 & 5.0 \textpm \, 2.9 & 16.6 \textpm \, 0.5 & 17.7 \textpm \, 3.1 & \textbf{27.3 \textpm \, 2.6} \\ [1ex] 
 \hline
\end{tabular}
    \caption{Macro F1 Scores with Symmetric Instance-Invariant Noisy Labels on CIFAR10.}
    \label{tab:noise_cifar10_sym}
\end{table}

\subsection{Detecting Incorrect Labels}

In addition to showing that ASIF confers an advantage when training in the face of noisy labels, we can also show that ASIF can help to detect which samples have bad labels.

To do so, periodically through the training we recorded the associated $\mathscr{L}_\text{cls}$ values for each sample and then ranked them by descending value. The top $N \times \eta$ samples were selected as 'probably false'. These samples' labels were then compared with the true labels to determine the dirty label picking balanced accuracy.

Our results indicate that utilising ASIF to pick out bad labels can confer as much as a 10\% advantage over the baseline CE loss. This is a significant improvement over both GCE and PHuber. In the easier scenario of Symmetric Instance-Invariant noise, ASIF shows a detection advantage of as much as 13\% over CE.

Please see Tables \ref{tab:label_picking_cifar10_inst} and \ref{tab:label_picking_cifar10_sym} for full results.

\begin{table}[]
    \centering
\begin{tabular}{||c c c c c||} 
 \hline
 $\eta$ & CE & GCE & PHuber & ASIF \\ [0.ex] 
 \hline\hline
 0.2 & 64.2 \textpm \, 0.8 & 65.0 \textpm \, 1.0 & 67.6 \textpm \, 1.1 & \textbf{70.0 \textpm \, 1.6} \\ 
 \hline
 0.4 & 73.4 \textpm \, 0.5 & 70.5 \textpm \, 1.4 & 72.6 \textpm \, 5.0 & \textbf{78.0 \textpm \, 1.6} \\ 
 \hline
 0.6 & 73.2 \textpm \, 1.6 & 67.7 \textpm \, 0.5 & 74.5 \textpm \, 0.6 & \textbf{79.7 \textpm \, 3.3} \\ 
 \hline
 0.7 & 69.7 \textpm \, 1.6 & 66.2 \textpm \, 1.1 & 64.5 \textpm \, 3.1 & \textbf{79.6 \textpm \, 2.0} \\ 
 \hline
 0.8 & 65.9 \textpm \, 0.7 & 60.7 \textpm \, 0.7 & 62.9 \textpm \, 3.6 & \textbf{71.2 \textpm \, 1.2} \\
 \hline
 0.9 & 58.1 \textpm \, 2.0 & 58.3 \textpm \, 0.9 & 61.1 \textpm \, 0.6 & \textbf{61.9 \textpm \, 3.0} \\ [1ex] 
 \hline
\end{tabular}
    \caption{F1 Score of Instance-Dependent Noisy Label Detection on CIFAR10}
    \label{tab:label_picking_cifar10_inst}
\end{table}

\begin{table}[]
    \centering
\begin{tabular}{||c c c c c||} 
 \hline
 $\eta$ & CE & GCE & PHuber & ASIF \\ [0.ex] 
 \hline\hline
 0.2 & 79.1 \textpm \, 3.1 & \textbf{86.7 \textpm \, 0.6} & 82.0 \textpm \, 2.0 & 85.6 \textpm \, 0.7 \\ 
 \hline
 0.4 & 75.9 \textpm \, 0.2 & 83.7 \textpm \, 0.3 & 84.6 \textpm \, 0.5 & \textbf{86.2 \textpm \, 1.3} \\ 
 \hline
 0.6 & 73.3 \textpm \, 0.9 & 77.6 \textpm \, 1.1 & 83.1 \textpm \, 1.4 & \textbf{84.1 \textpm \, 0.4} \\ 
 \hline
 0.7 & 67.1 \textpm \, 1.6 & 73.2 \textpm \, 1.1 & \textbf{79.7 \textpm \, 3.2} & 79.5 \textpm \, 1.3 \\ 
 \hline
 0.8 & 60.7 \textpm \, 0.4 & 65.7 \textpm \, 0.7 & 71.7 \textpm \, 0.6 & \textbf{73.2 \textpm \, 2.0} \\
 \hline
 0.9 & 52.2 \textpm \, 2.2 & 54.8 \textpm \, 0.9 & 59.4 \textpm \, 2.0 & \textbf{62.9 \textpm \, 0.9} \\ [1ex] 
 \hline
\end{tabular}
    \caption{F1 Score of Symmetric Instance-Invariant Noisy Label Detection on CIFAR10}
    \label{tab:label_picking_cifar10_sym}
\end{table}

\subsection{Analysis of the Features Learned With ASIF}
\label{FeatureAnalysis}

We conducted an analysis to gain a better understanding of the effect ASIF has on the features learned. First, we address the degree to which ASIF inhibits memorisation of features. The theory behind ASIF is that we penalise the learning of features that can be used to identify specific instances of data, while allowing the learning of features that are necessary to differentiate between classes of data. We tested this by training a single layer classifier on the feature vectors obtained from the frozen pre-trained feature extractor $\Psi(x)$ until performance stopped improving. The goal was correctly identifying each individual sample. If ASIF performed as theorised, the best loss obtained would be worse for ASIF than it would be for the baseline. The results, shown in Table \ref{tab:feature_vector_ids_losses}, confirm this supposition.

\begin{table}[]
    \centering
\begin{tabular}{||c c c||} 
 \hline
 Dataset & CE & ASIF \\ [0.ex] 
 \hline\hline
 CIFAR10 & 0.005 \textpm \, 0.008 & 0.782 \textpm \, 0.159 \\ 
 \hline
 Fashion-MNIST & 1.059 \textpm \, 0.332 & 8.307 \textpm \, 0.202 \\ [1ex] 
 \hline
\end{tabular}
    \caption{Loss Obtained identifying feature vectors}
    \label{tab:feature_vector_ids_losses}
\end{table}

Second, we turn our attention to feature dis-entanglement. There is a tension behind the two goals of class discrimination and instance discrimination, since features can be used toward both ends. We hypothesize that the result of this struggle would be that only the fewest, most useful features would be selected for classification. This would act to dis-entangle class-wise and identity-wise features, with class-variant features confined to a small subspace and the rest of the features showing no significant statistical differences between classes.

We set out to investigate the distribution of the features selected by ASIF. To do so, we performed an experiment. We trained a fresh linear classifier $h$ on the feature vectors obtained from the frozen pre-trained feature extractor $\Psi(x)$. We then used the absolute value of the linear weights $h_\Theta$ to ascertain which feature dimensions were least important to the classification and removed those from the vectors, before retraining another fresh classifier on the resulting vectors. We performed this process repeatedly, reducing the vector size with each iteration until the resulting vectors had only five dimensions. With each iteration, we recorded the best classification accuracy obtained.

\begin{figure}
\centering
    \includegraphics[width=3.2in]{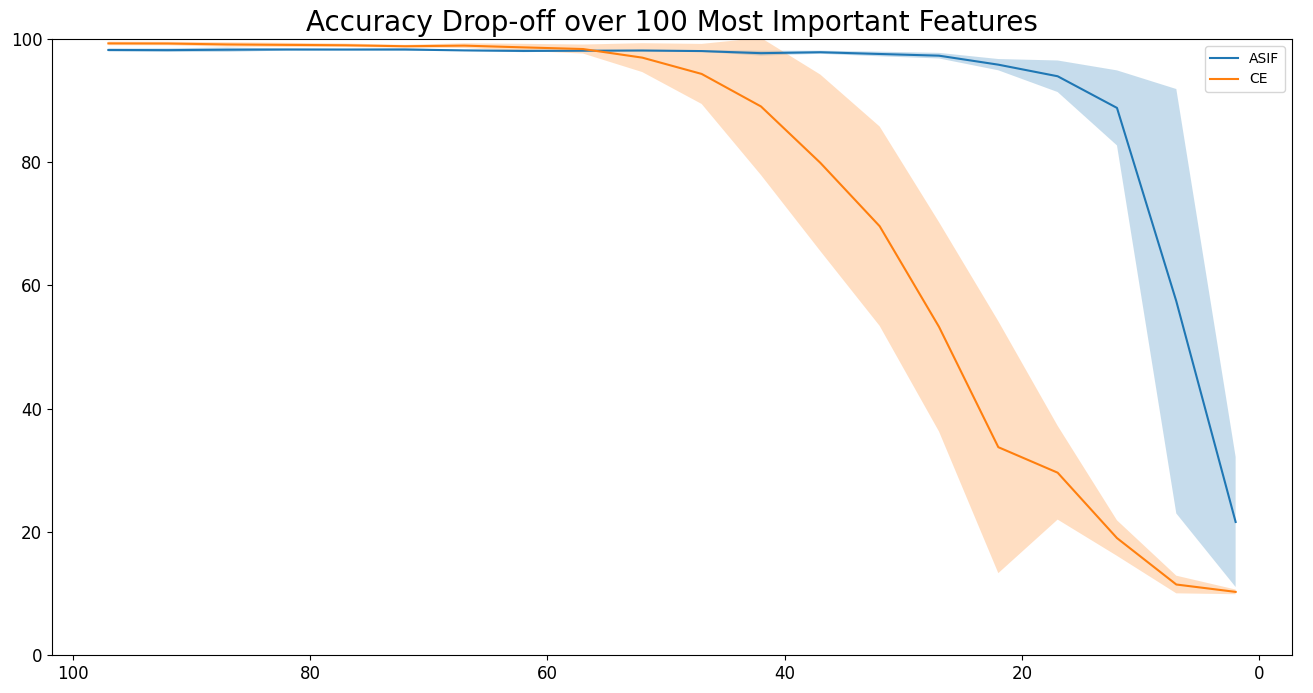}
    \caption{The accuracy obtained when training a single-layer classifier on the output of feature extractors trained with Cross Entropy Loss (Red) and ASIF (Blue) using between 1 and 100 of the most important features.}
    \label{fig:clustering_accuracy_vs_features}
\end{figure}

Figure \ref{fig:clustering_accuracy_vs_features} shows the results obtained from this experiment. In all cases, the majority of the features played no important role in classification. However, as the least important features continued to be removed, the Cross Entropy baseline became the first to show a reduced accuracy when there were around 55 feature dimensions remaining. 

ASIF, on the other hand, does not seriously lose accuracy until we are down to the ten most important feature dimensions. This feature compression is desirable as it has been linked to generalization \cite{tishby2015deep, shwartzziv2017opening}.

\section{Conclusions and Future Work}
\label{Conclusions}

We have introduced ASIF, which differentiates between Class-wise and Identity-wise Discriminative Features, promoting the former whilst suppressing the latter. Through experimentation, we have shown that ASIF has a regularising effect that can reduce overfitting to individual samples and increase robustness against label noise. We have shown that ASIF can result in accuracies that are better than standard training results, especially in the high-noise domain. Moreover, we have also proposed the DGR method which allows for a gradient reversal layer without tune-able hyper-parameters.

We have shown that the use of ASIF results in the selection of far fewer class-variant features. This feature of ASIF-trained feature extractors has great potential for use in Data Privacy and Anonymization applications, or anywhere else that autoencoders are useful.

Moreover, when training on the full dataset, the selected features are likely to be domain invariant and lead to much better generalisation. Extending the Identification task to the unsupervised realm may encourage more domain invariant feature selection and seems like a fruitful direction for future research.

ASIF has limitations when it comes to scaling up. Because its Identifier module has a linear layer that specifies individual instances in a fixed dataset, the size of that dataset has a practical upper limit. Because each class has its own set of weights, this memory constraint also places a limit on the number of classes trained. Moreover, while suppressing identity features has been shown to be helpful in the settings laid out in this paper, they are crucial in certain other tasks - such as face recognition. This limits ASIF's applicability in some areas. Despite these limitations, differentiating between Class-wise and Identity-wise Features is nonetheless a novel approach.

\section*{Acknowledgment}

This work was supported with supercomputing resources provided by the Phoenix HPC service at the University of Adelaide.

\bibliography{references.bib}
\bibliographystyle{ieeetr}

\newpage
\appendices

\section{CIFAR10 ASIF Configurations}
\label{appendix:cifar10_asif_config}

Table \ref{tab:cifar10_asif_experiment_configs} shows the configurations used to produce the CIFAR10 ASIF results.

\begin{table}[]
    \centering
\begin{tabular}{||c c c c c||} 
 \hline
 Noise & $\eta$ & N & LR & $\lambda_\text{if}$ \\ [0.ex] 
 \hline\hline
 Instance & 0.2 & 50k & 0.0001 & 0.01 \\ 
 \hline
 Instance & 0.4 & 50k & 0.0001 & 0.1 \\ 
 \hline
 Instance & 0.6 & 50k & 0.0001 & 0.1 \\ 
 \hline
 Instance & 0.7 & 50k & 0.0001 & 0.1 \\ 
 \hline
 Instance & 0.8 & 50k & 0.0001 & 0.1 \\ 
 \hline
 Instance & 0.9 & 50k & 0.0001 & 0.1 \\ 
 \hline
 Symmetric & 0.2 & 50k & 0.0001 & 10.0 \\ 
 \hline
 Symmetric & 0.4 & 50k & 0.0001 & 100.0 \\ 
 \hline
 Symmetric & 0.6 & 50k & 0.0001 & 1.0 \\ 
 \hline
 Symmetric & 0.7 & 50k & 0.0001 & 1.0 \\ 
 \hline
 Symmetric & 0.8 & 50k & 0.001 & 1.0 \\ 
 \hline
 Symmetric & 0.9 & 50k & 0.001 & 10.0 \\ 
 \hline
 None & 0 & 10k & 0.001 & 1.0 \\ 
 \hline
 None & 0 & 20k & 0.0001 & 1.0 \\ 
 \hline
 None & 0 & 30k & 0.0001 & 1.0 \\ 
 \hline
 None & 0 & 40k & 0.0001 & 1.0 \\ 
 \hline
 None & 0 & 50K & 0.001 & 1.0 \\  [1ex] 
 \hline
\end{tabular}
    \caption{CIFAR10 ASIF Experimental Configurations.}
    \label{tab:cifar10_asif_experiment_configs}
\end{table}

\section{Fashion-MNIST ASIF Configurations}
\label{appendix:cifar10_fashion_mnist_config}

Table \ref{tab:fashion_mnist_asif_experiment_configs} shows the configurations used to produce the Fashion-MNIST ASIF results.

\begin{table}[]
    \centering
\begin{tabular}{||c c c c c||} 
 \hline
 Noise & $\eta$ & N & LR & $\lambda_\text{if}$ \\ [0.ex] 
 \hline\hline
 Instance & 0.2 & 60k & 0.001 & 0.001 \\ 
 \hline
 Instance & 0.4 & 60k & 0.0001 & 0.1 \\ 
 \hline
 Instance & 0.6 & 60k & 0.001 & 0.001 \\ 
 \hline
 Instance & 0.7 & 60k & 0.001 & 0.001 \\ 
 \hline
 Instance & 0.8 & 60k & 0.001 & 1.0 \\ 
 \hline
 Instance & 0.9 & 60k & 0.001 & 0.001 \\ 
 \hline
 Symmetric & 0.2 & 60k & 0.0001 & 0.001 \\ 
 \hline
 Symmetric & 0.4 & 60k & 0.001 & 0.001 \\ 
 \hline
 Symmetric & 0.6 & 60k & 0.001 & 0.001 \\ 
 \hline
 Symmetric & 0.7 & 60k & 0.001 & 1.0 \\ 
 \hline
 Symmetric & 0.8 & 60k & 0.0001 & 10.0 \\ 
 \hline
 Symmetric & 0.9 & 60k & 0.0001 & 0.1 \\ 
 \hline
 None & 0 & 10k & 0.001 & 0.001 \\ 
 \hline
 None & 0 & 20k & 0.001 & 0.001 \\ 
 \hline
 None & 0 & 30k & 0.001 & 0.001 \\ 
 \hline
 None & 0 & 40k & 0.001 & 0.001 \\ 
 \hline
 None & 0 & 60K & 0.001 & 0.01 \\  [1ex] 
 \hline
\end{tabular}
    \caption{Fashion-MNIST ASIF Experimental Configurations.}
    \label{tab:fashion_mnist_asif_experiment_configs}
\end{table}

\section{Fashion-MNIST Detailed Results}
\label{appendix:fashion_mnist_detailed_results}

Experiments run on CIFAR10 were also performed on the Fashion-MNIST dataset to test repeatability of the results. Unlike in the case of CIFAR10, tests were only run on Cross Entropy and ASIF, not on GCE or PHuber.

Results are shown here:
\begin{itemize}
\item Reduced Datasets: Table \ref{tab:small_set_fashion_mnist_results}
\item Symmetrical Instance-Invariant Noise: Table \ref{tab:noise_fashion_mnist_sym}
\item Instance-Dependent Noise: Table \ref{tab:noise_fashion_mnist_inst}
\end{itemize}

\begin{table}[]
    \centering
\begin{tabular}{||c c c||} 
 \hline
 $N$ & CE & ASIF \\ [0.ex] 
 \hline\hline
 10k & 89.1 \textpm \, 0.2 & \textbf{89.6 \textpm \, 0.1} \\ 
 \hline
 20k & 90.5 \textpm \, 0.1 & \textbf{91.2 \textpm \, 0.3} \\ 
 \hline
 30k & 91.4 \textpm \, 0.1 & \textbf{91.9 \textpm \, 0.1} \\ 
 \hline
 40k & 92.0 \textpm \, 0.1 & \textbf{92.5 \textpm \, 0.3} \\ 
 \hline
 60k & 93.0 \textpm \, 0.0 & \textbf{93.1 \textpm \, 0.2} \\ [1ex] 
 \hline
\end{tabular}
    \caption{Macro F1 scores when training on reduced training sets on Fashion-MNIST.}
    \label{tab:small_set_fashion_mnist_results}
\end{table}

\begin{table}[]
    \centering
\begin{tabular}{||c c c||} 
 \hline
 $\eta$ & CE & ASIF \\ [0.ex] 
 \hline\hline
 0 & 93.0 \textpm \, 0.0 & \textbf{93.1 \textpm \, 0.2} \\ 
 \hline
 0.2 & 88.1 \textpm \, 0.0 & \textbf{90.3 \textpm \, 0.4} \\ 
 \hline
 0.4 & 83.7 \textpm \, 0.1 & \textbf{88.8 \textpm \, 1.1} \\ 
 \hline
 0.6 & 74.9 \textpm \, 1.1 & \textbf{86.9 \textpm \, 0.8} \\ 
 \hline
 0.7 & 66.7 \textpm \, 2.2 & \textbf{85.3 \textpm \, 0.8} \\ 
 \hline
 0.8 & 57.3 \textpm \, 4.8 & \textbf{81.4 \textpm \, 2.3} \\
 \hline
 0.9 & 19.4 \textpm \, 3.5 & \textbf{73.2 \textpm \, 0.4} \\ [1ex] 
 \hline
\end{tabular}
    \caption{Macro F1 Scores when training on Fashion-MNIST with Symmetric Instance-Invariant Noisy Labels.}
    \label{tab:noise_fashion_mnist_sym}
\end{table}

\begin{table}[]
    \centering
\begin{tabular}{||c c c||} 
 \hline
 $\eta$ & CE & ASIF \\ [0.ex] 
 \hline\hline
 0 & 93.0 \textpm \, 0.0 & \textbf{93.1 \textpm \, 0.2} \\ 
 \hline
 0.2 & \textbf{85.6 \textpm \, 0.2} & 82.5 \textpm \, 4.3
 \\ 
 \hline
 0.4 & \textbf{74.7 \textpm \, 0.6} & 72.4 \textpm \, 1.9 \\ 
 \hline
 0.6 & 53.4 \textpm \, 0.6 & \textbf{58.3 \textpm \, 1.5} \\ 
 \hline
 0.7 & 43.1 \textpm \, 1.0 & \textbf{44.8 \textpm \, 1.6} \\ 
 \hline
 0.8 & 30.5 \textpm \, 0.6 & \textbf{34.3 \textpm \, 0.8} \\
 \hline
 0.9 & 21.4 \textpm \, 1.6 & \textbf{24.7 \textpm \, 2.4} \\ [1ex] 
 \hline
\end{tabular}
    \caption{Macro F1 Scores when training on Fashion-MNIST with Instance-Dependent Noisy Labels.}
    \label{tab:noise_fashion_mnist_inst}
\end{table}

\end{document}